# Automatic Pulmonary Nodule Detection in CT Scans Using Convolutional Neural Networks Based on Maximum Intensity Projection

Sunyi Zheng, Jiapan Guo*, Xiaonan Cui, Raymond N. J. Veldhuis, Matthijs Oudkerk, and Peter M.A. van Ooijen

*Abstract*—Accurate pulmonary nodule detection is a crucial step in lung cancer screening. Computer-aided detection (CAD) systems are not routinely used by radiologists for pulmonary nodule detection in clinical practice despite their potential benefits. Maximum intensity projection (MIP) images improve the detection of pulmonary nodules in radiological evaluation with computed tomography (CT) scans. Inspired by the clinical methodology of radiologists, we aim to explore the feasibility of applying MIP images to improve the effectiveness of automatic lung nodule detection using convolutional neural networks (CNNs). We propose a CNN-based approach that takes MIP images of different slab thicknesses (5 mm, 10 mm, 15 mm) and 1 mm axial section slices as input. Such an approach augments the two-dimensional (2-D) CT slice images with more representative spatial information that helps discriminate nodules from vessels through their morphologies. Our proposed method achieves sensitivity of 92.67% with 1 false positive per scan and sensitivity of 94.19% with 2 false positives per scan for lung nodule detection on 888 scans in the LIDC-IDRI dataset. The use of thick MIP images helps the detection of small pulmonary nodules (3 mm-10 mm) and results in fewer false positives. Experimental results show that utilizing MIP images can increase the sensitivity and lower the number of false positives, which demonstrates the effectiveness and significance of the proposed MIP-based CNNs framework for automatic pulmonary nodule detection in CT scans. The proposed method also shows the potential that CNNs could gain benefits for nodule detection by combining the clinical procedure.

(Corresponding author: Jiapan Guo.)
S. Zheng and P.M.A. van Ooijen are with the University of Groningen, University Medical Center Groningen, Department of Radiation Oncology, 9713 GZ Groningen, The Netherlands (e-mail: s.zheng@umcg.nl; p.m.a.van.ooijen@umcg.nl).
J. Guo is with the University of Groningen, University Medical Center Groningen, Department of Radiotherapy, 9713 GZ Groningen, The Netherlands (e-mail: j.guo@umcg.nl).
X. Cui is with the University of Groningen, University Medical Center Groningen, Department of Radiology, 9713 GZ Groningen, The Netherlands and also with the Tianjin Medical University Cancer Institute and Hospital, National Clinical Research Centre of Cancer, Department of Radiology, 300060 Tianjin, China (e-mail: x.cui@umcg.nl).
R. N. J. Veldhuis is with the University of Twente, 7500 AE Enschede, The Netherlands (e-mail: r.n.j.veldhuis@utwente.nl).
M. Oudkerk is with the Univeristy of Groningen, 9713 AV Groningen, The Netherlands (e-mail: m.oudkerk@umcg.nl)

## I. INTRODUCTION

LUNG cancer, as one of the most severe cancers with high incidence, has a devastating effect on human lives [1]. It has been predicted to be one of the greatest single cause of mortality among the European population in 2019 [2]. The disease can effectively be treated with radiotherapy and chemotherapy. However, individuals who are diagnosed with lung cancer only have a 16% five-year survival rate [3]. Early detection of lung cancer is a crucial step since it could improve chances of survival [4].

With continuously updated technology, computer-aided detection (CAD) plays an increasingly important role to assist radiologists in staging lung cancer tumors. It also helps to improve the accuracy of lung nodule detection as well as to reduce the number of missed nodules and misdiagnosis [5]. Nodule candidate detection and false positive reduction are two essential parts in a well-performing pulmonary CAD system. Nodule candidate detection tries to find as many nodule candidates as possible, whereas false positive reduction as a necessary follow-up procedure aims to eliminate wrong findings. In the nodule candidate detection stage, the system attempts to include all the potential candidates. But it could also detect numerous false positives, which results in interference for the diagnosis. In the false positive reduction stage, although some CAD systems can achieve a low false positive rate with a high sensitivity, it can still miss a few nodules. The difficulty of CAD systems for detection is because of the variety of nodules. Specifically, pulmonary nodules have complex features, such as size, shapes, margin information and calcification patterns. These features are also used to diagnose nodules [6].

To provide benefits for early diagnosis, a large number of researchers have developed CAD systems for nodule detection. Gurcan et al. [7] proposed rule-based classifiers to differentiate lung nodules from other similar structures by utilizing 2-D and 3-D features, followed by a linear discriminant analysis to reduce false positives. In another approach, Messay et al. [8] combined the multiple intensity thresholding approaches and morphological opening to detect nodule candidates. Then, they used the Fisher Linear Discriminant (FLD) classifier and a quadratic classifier to discriminate between false predictions and true candidates. In the effort to detect nodules by their features, Wook-Jin Choi and Tae-Sun Choi [9] proposed a three-dimensional shape-based feature descriptor for nodule detection and a support vector machine-based classifier to refine the detected candidates. In recent years, deep learning algorithms have shown outstanding performance in computer science for object detection [10], natural scene recognition [11], medical image analysis [12], etc. An increasing number of newer approaches, relying on deep learning techniques, has been applied for lung nodule detection. For example, Ypsilantis et al. [13] exploited a recurrent neural network (RNN) to detect pulmonary nodules, in order to increase the sensitivity without increasing the false positive rate. In a different method, Setio et al. [14] used 2-D multi-view convolutional neural networks and a dedicated fusion method to combine the classification results from different views. To reduce the number of false positives, Dou et al. [15] tried multilevel contextual 3-D convolutional neural networks (CNNs) by utilizing different sizes of cubes as the input. In addition, Jiang et al. [16] attempted to cut out multigroup patches from the lung parenchyma. They implemented four different 2-D convolutional networks for the nodule detection with various patch sizes. Another approach is that Narayanan et al. [17] utilized CT images with different slice thicknesses as training data to detect candidates. To include all the potential nodule candidates, Zhang et al. [18] proposed a method, using multi-scale LoG filters to localize nodules. Further, a densely dilated 3-D deep convolutional neural network was applied to reduce the number of false positives.

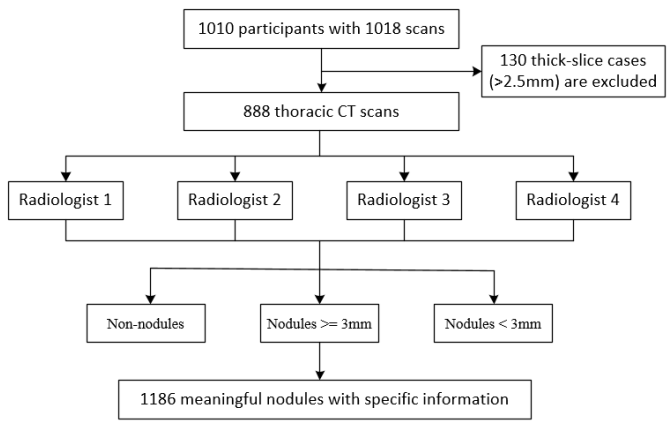

Fig. 1. The selection process of meaningful nodules in the LUNA 16 challenge.

Although these CAD systems showed high efficiency and benefits in lung nodule detection, only a few studies developed approaches that take the routine workflow of the radiologists into consideration. In clinical practice, radiologists first take a quick look at the maximum intensity projection (MIP) images in order to roughly locate the nodule candidates for further examinations on specific slices. MIP [19] is a postprocessing method that projects 3-D voxels with maximum intensity to the plane of projection. It is widely used for the detection of lung nodules in the nodule screening since it enhances the visualization of nodules in comparison to the presence of bronchi and vasculature. Inspired by such an initial procedure, we propose a novel CAD system based on deep learning, which we call a MIP-based convolutional neural networks approach. The purpose of this study is to explore whether the images based on MIP could help convolutional neural networks to increase accuracy in automatic detection of lung nodules.

The main contributions of this paper are as follows. (1) Following clinical procedures, we took into account the workflow of lung cancer screening by radiologists who used MIP images during the nodule screening stage. (2) We proposed a multi-views MIP-based framework for lung nodule detection that adds spatial information into 2-D images to discriminate between nodules and vessels. (3) We applied multiple deep CNNs to reduce false positive candidates and fused the final predictions. This provides complementary information and leads to more accurate and robust performance of the system. (4) We provided a clinical-methodology-based lung nodule detection system that is more easily interpretable for radiologists. (5) We proved the feasibility of combining the clinical screening method and CNNs to improve the detection of lung nodules.

## II. MATERIALS

### A. Database

The Lung Image Database Consortium and Image Database Resource Initiative (LIDC/IDRI) [20] was an established repository of computed tomography scans to facilitate computer-aided systems on the assessment of lung nodule detection, classification and quantification. The LIDC/IDRI database consists of 1018 thoracic CT scans with the corresponding nodule annotations. Based on this database, the Lung Nodule Analysis 2016 (LUNA16) challenge was launched as a benchmarking to allow large-scale evaluation of automatic nodule detection algorithms. In this dataset, the image slice thickness varied from 0.6 mm to 5.0 mm. The competition removed scans with the slice thickness larger than 2.5 mm since it was not recommended for clinical nodule screening [21, 22]. Therefore, in total 888 scans remained. They acquired annotated nodule information after the scans were viewed by four experienced radiologists in a two-stage reading procedure. More specifically, every radiologist annotated nodules independently in the first period, whereafter they repeated the analysis separately based on the unblinded four results from the first stage. There were three nodule categorizations which were non-nodules, nodules with a size less than 3 mm in diameter and nodules with a size of equal to or larger than 3 mm in diameter. The non-nodule and small nodule categorizations were not taken into consideration since they lack clinical relevance [23]. Hence, only the lung nodules equal to or larger than 3 mm were manually segmented. In addition, the nodules accepted by at least 3 out of 4 radiologists were used as the reference standard. In this way, 1186 valid nodules remained for the experiments. The procedure of nodule selection is displayed in Fig. 1. In this work, we use the diameter of each nodule provided in the LUNA16 competition to generate a bounding box as the label for the detection of nodules. For detailed information about LUNA16 see [24].

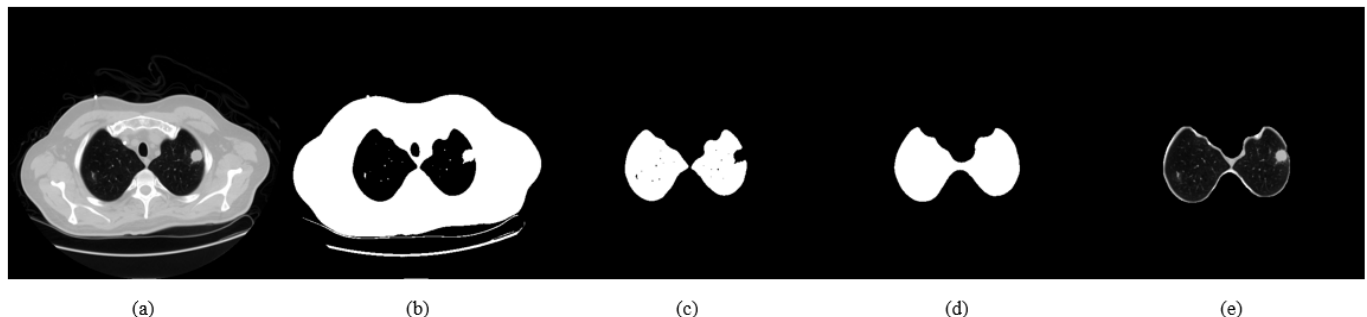

Fig. 2. The procedures of lung parenchyma segmentation. (a) A raw thoracic CT image; (b) The mask of labeled chest; (c) The mask after the removal of irrelevant objects; (d) The mask of patched lung parenchyma; (e) The image of segmented lung parenchyma.

## III. METHODS

Our proposed system for nodule detection consists of two stages, namely nodule candidate detection and false positive reduction. At the nodule candidate detection stage, we first segment the lung parenchyma and then apply four 2-D convolutional neural networks, each of which takes the MIP images with a certain slab parameter as inputs. We then merge the detected nodules from these four CNNs as nodule candidates. During the second stage, we use cubic image patches which are centered on the nodule candidates for training. We ensemble 3-D convolutional neural networks to classify them as real nodules or not. Details of the CAD system are given in the following sections.

### A. Lung Parenchyma Segmentation

To exclude irrelevant regions, such as clothes, machine objects, tissues, spines, or ribs, we employ a scheme to automatically segment the lung parenchyma in thoracic CT scans. The processing steps of segmenting lung parenchyma are shown in Fig. 2. Since the data set is collected from various CT scanners in the LIDC/IDRI dataset, we set the window level from -1000 HU to 400 HU and normalize images to the range between 0 and 1. We use the mean value of the whole image as a threshold to divide the chest into the outside and inside regions for a rough classification, as shown in Fig. 2(b). Then we remove isolated pixels which are connected to the white label border. We also adopt a 4-connected neighborhood operator to eliminate these unrelated tissues as well as the noise from the CT detector. The two largest connected black components are selected as the internal chest region. This gives a mask of the internal chest and removes the outer unimportant area of the chest, as shown in Fig. 2(c). In order to avoid missing wall-attached lesions that might be nodules, we keep more boundary information by binary morphology operations, i.e., closing and dilation. However, there are some small noisy areas, such as graininess, vessels and tissues, which appear in the processed image. The morphology method called "binary fill holes" [25] is applied. This approach eliminates graininess and vessels that incurred into the lung cavity by the removal of radio-opaque tissue. The complete mask can be seen in Fig. 2(d) and the well-segmented lung parenchyma is shown in Fig. 2(e).

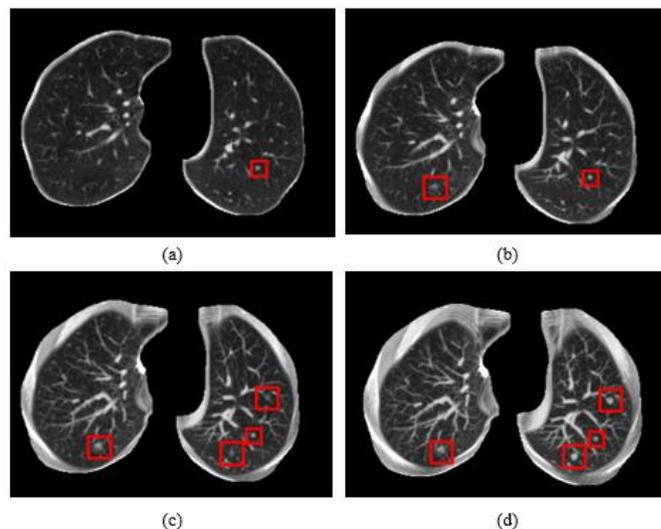

Fig. 3. Examples of maximum intensity projection (MIP) images with diverse slab thicknesses. (a) A 1 mm axial section slice. (b-d) MIP images under the slab thickness of 5 mm, 10 mm and 15 mm, respectively. Each nodule is highlighted by a red bounding box.

### B. Maximum Intensity Projection

In clinical practice, radiologists go through several 1 mm axial section slices to differentiate between suddenly appeared nodules and vessels, since nodules can be easily confused with vessels in a single slice. Moreover, maximum intensity projection images are also used routinely by radiologists to improve detection of pulmonary nodules, especially small nodules. MIP images are the superposition of maximum grey values at each coordinate from a stack of consecutive slices. Such a combined image shows morphological structures of isolated nodules and continuous vessels. In our work, we use MIP images as the input for convolutional neural networks to detect nodule candidates. To normalize the size of images, we rescale each image to a 1 mm slice spacing in the z-direction and keep the original spacing in the other two relatively vertical planes (i.e., x and y directions). For a wide range of nodule diameters, this step creates 1 mm axial section slices and MIP images with slab thickness equals to 5 mm, 10 mm, 15 mm from each scan. We choose these slab thicknesses because clinically they show significant improvements in nodule detection [26-29]. Specifically, they enhance textural information and prevent missing details of the nodule which are occluded by vessels or lung tissues. Examples of MIP images with various slab parameters are shown in Fig. 3. These images are from the same patient scan. The isolated nodules and consecutive vessels are better visible in MIP images than those in the 1 mm axial section slices.

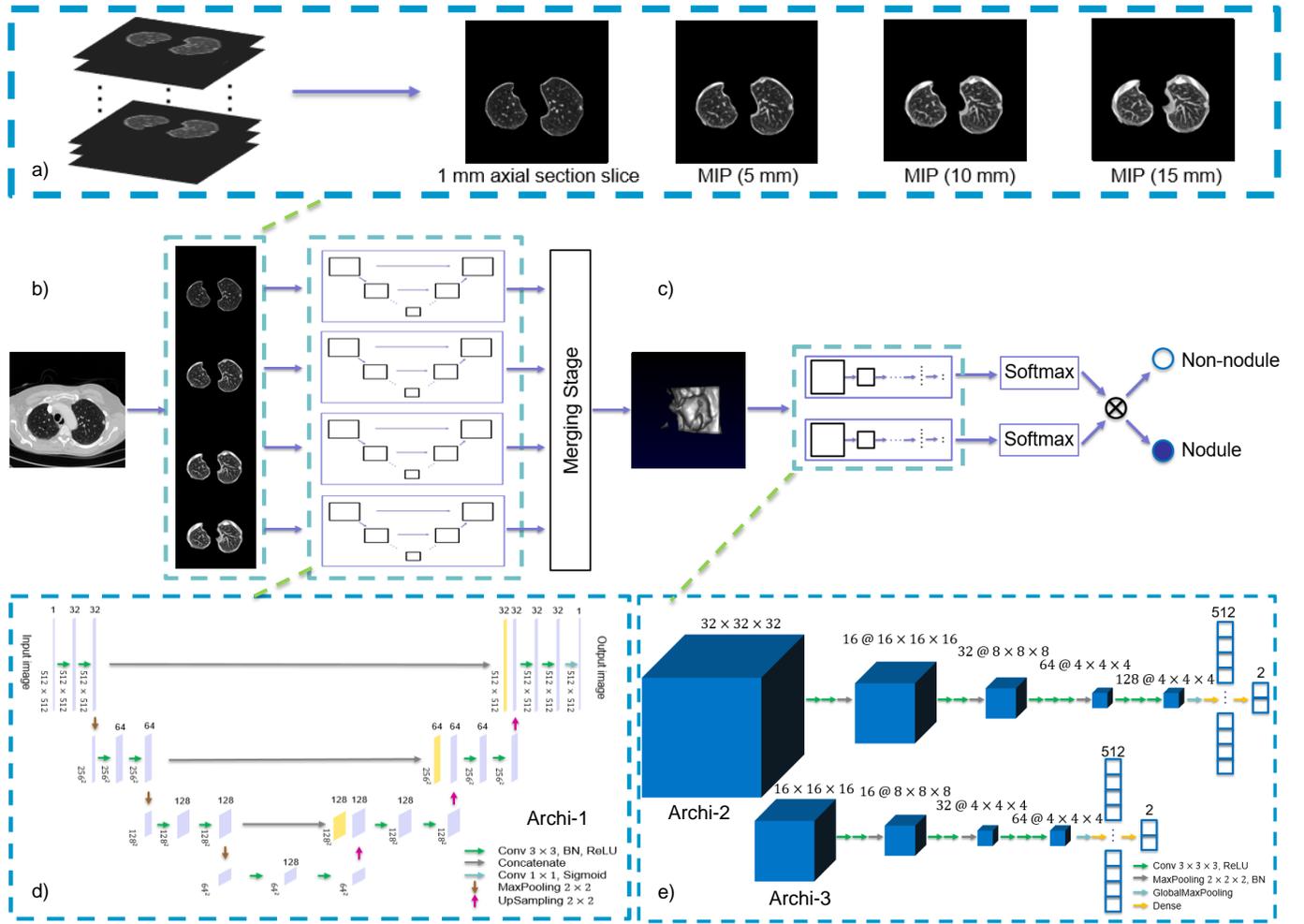

Fig. 4. The overview of the designed computer-aided detection system. (a) Examples of input images for architecture 1 (Archi-1). Those 1 mm axial section slices and maximum intensity projection (MIP) images with varied slab thicknesses (5 mm, 10 mm and 15 mm) are generated as inputs. (b) The scheme for nodule candidate detection in CT scans. Every scan is checked by networks in four streams and all the potential candidates are fused at the merging stage. (c) The framework for false positive reduction. After soft-max layer, results are merged to give a probability of being a nodule to each location. (d) The structure of architecture 1 for finding potential candidates. (e) The structures of architecture 2 and 3 (Archi-2 and Archi-3) with cubic patches of various size for the reduction of false positives.

### C. MIP-based Convolutional Networks and Model Fusion

Fig. 4(b) illustrates the proposed MIP-based 2-D CNN framework with different slab thicknesses for automatic lung nodule detection. The nodule detection stage is constructed by combining four streams of the 2-D convolutional neural network. Each stream acquires different contextual information and the discrimination of nodules from vessels through their morphologies based on the images in a specific slab thickness.

The 2-D CNN architecture is shown in Fig. 4(d). It is based on the U-net [30], which consists of an encoding and a decoding part. The inputs of the network are MIP images of size $512 \times 512$ pixels with different slab thicknesses (Fig. 4(a)). In the encoding part, there is a repeated application of two convolutions with kernel sizes of $3 \times 3$, two batch normalization layers, two activation layers and a max-pooling layer. The convolution layer starts with a 32-component feature map, and the number of feature components doubles after every application. Each convolutional layer is followed by a batch normalization layer and a rectified linear unit (ReLU) as the activation function [31]. After the second activation function in every application, there is a $2 \times 2$ max pooling operation with a stride of 2. In the decoding part, the application starts from a $2 \times 2$ up-sampling layer followed by concatenation with the corresponding feature map from the encoding period, two $3 \times 3$ convolutional layers, two batch normalization layers and two ReLU activation layer. Specifically, a batch normalization layer and a ReLU activation function come after each convolutional layer. The convolution in the decoding part starts with a 128-component feature map and the number of feature components halves after every application. The last layer is a

TABLE I
PERFORMANCE OF THE SYSTEM AT THE NODULE CANDIDATE DETECTION STAGE

Total number of scans: 888
Total number of nodules: 1186 (3-10 mm: 905, 10-20 mm: 231, ≥20 mm: 50)

| Stream | The slab thickness of the training data | Number of detected nodules (3-10 mm) | Number of detected nodules (10-20 mm) | Number of detected nodules (≥20 mm) | Total number of detected nodules | Sensitivity (%) | False Positives (FPs) | FPs per scan |
|---|---|---|---|---|---|---|---|---|
| Stream 1 | 1 mm | 719 | 213 | 50 | 982 | 82.80 | 12,940 | 14.57 |
| Stream 2 | 5 mm | 774 | 218 | 50 | 1,042 | 87.86 | 9,792 | 11.03 |
| Stream 3 | 10 mm | 801 | 216 | 50 | 1,067 | 89.97 | 6,895 | 7.76 |
| Stream 4 | 15 mm | 787 | 215 | 50 | 1,052 | 88.70 | 5,602 | 6.31 |
| Fusion | - | 856 | 225 | 50 | 1,131 | 95.36 | 16,985 | 19.13 |

$1 \times 1$ convolution that is followed by a sigmoid activation function. In total, the network has 18 convolutional layers.

In the stage of merging results, we combine a contour retrieval mode and an approximation method [32] to extract and refine the predicted candidates' label information (i.e., coordinates and bounding box widths). In order to detect as many nodules as possible, we take all the potential candidates with regular-shaped predicted labels into consideration. We also notice that there are some candidates of which the centers are too close to each other. We used a distance ratio of 1.1 to consider whether two candidates are one finding or two individual findings. The distance ratio is the distance between the centers of two detected candidates on the same slice divided by the predicted side of the bounding box from the larger candidate. It is determined by the grid search experiments on a pilot study on one-fold of the dataset. Increasing this ratio leads to the loss of close nodules, although the number of false positives decrease. Two separate nodules should not be too close to each other. Besides, some different potential candidates may have the same location in the 2-D plane. These cases are differentiated through their slice numbers. Finally, we combine the results from each stream to obtain the fused prediction.

### D. False Positive Reduction

The proposed framework for classification between nodule and non-nodule is illustrated in Fig. 4(c). We train the 3-D CNN based on the merged candidates from the candidate detection stage to estimate the probability of being a nodule. We analyze the size from the potential candidates and most of the diameters are less than 16 pixels. Thus, we choose 16 pixels and 32 pixels as the cube sizes to differentiate nodules and non-nodules.

Fig. 4(e) shows the 3-D CNN scheme which includes Archi-2 and Archi-3 for discrimination between nodules and irrelevant findings. It is based on the VGG-net [33], which gives good performance on image classification and is a commonly used deep learning network for feature extraction. Based on the analysis of the size of nodules, we use cubic patches with volume sizes of $32 \times 32 \times 32$ and $16 \times 16 \times 16$ pixels as inputs to the networks. The Archi-3 consists of 12 layers, using the $16 \times 16 \times 16$ patches as inputs. There is a repeated application of layers which have two convolutional layers with 16 kernels of size $3 \times 3 \times 3$, a max-pooling layer and a batch normalization layer. This architecture contains three applications and the number of kernels doubles in every application. After all the applications, there is a global max-pooling layer which is used to retain the most important features in global parameters. The final part is two dense layers. Archi-2 has a structure similar to that of Archi-3. However, due to the larger patch size, this architecture includes three more convolutional layers, a max-poling layer and a batch normalization layer. In addition, the maximum number of kernels is 128.

### E. Training Process

In the nodule candidate detection stage, the whole dataset is equally split into 10 subsets in the LUNA16 competition. We perform 10-fold cross-validation to evaluate the performance of the system. For each fold, we use 63% of the dataset for training, 27% of the dataset for validation, and 10% of the dataset for testing. The best parameters are found in a pilot study on a smaller dataset. The batch size is set to 5 due to the maximum memory of the GPU and the depth of CNNs. The learning rate ranges from $10^{-3}$ to $10^{-7}$ with a reduced factor of 0.01 to learn features gradually and carefully. We use early stopping with a patience of 10 epochs to avoid overfitting. This study uses the dice score coefficient to calculate the loss value:

$$\text{Loss} = 1 - \frac{2|X \cap Y|}{|X| + |Y|} \quad (1)$$

Where $X$ and $Y$ are the predicted image and the ground truth image, respectively. The initializer in the convolution layer is He initialization[34] which has better performance for layers with ReLU activation. The He Initialization function is shown as follows:

$$y_l = W_l x_l + b_l \quad (2)$$

$$\frac{1}{2} n_l Var[w_l] = 1, \quad \forall l, \quad (3)$$

Where $W$ is a $d \times n$ matrix, $b$ is the vector of biases, $y$ is the response value at one pixel and $x$ is a $k^2 c$ vector which represents $k \times k$ pixels in $c$ input channels. The number of filters is $d$ and $l$ is used to index a layer. In equation (3), $w_l$ is a random variable in $W_l$ and $n$ equals to $k^2 c$. We use Adam stochastic optimization in the backpropagation [35]. We also augment the training data to improve the performance of the system. Image patches including nodules are translated for 30 voxels along each axis and they are rotated at angles of 0 degrees, 90 degrees, 180 degrees and 270 degrees randomly. Flip is added to transform images vertically, horizontally or both.

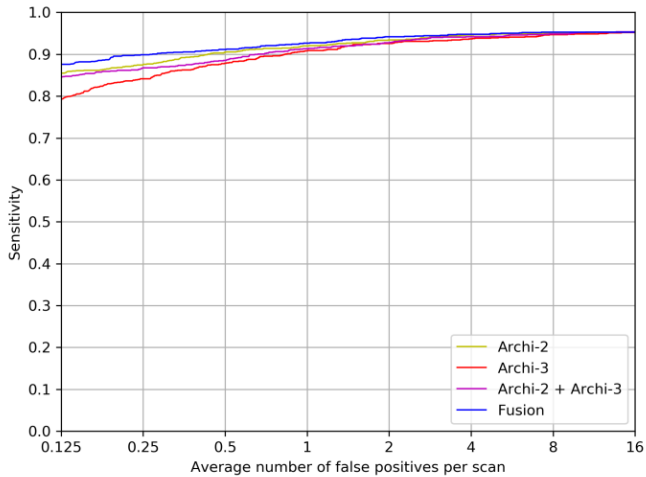

Fig. 5. Free-response receiver operating characteristic (FROC) curves of various architectures and the fusion results of the computer-aided detection system on the LIDC/IDRI database.

TABLE II
PERFORMANCE COMPARISON AT THE NODULE CANDIDATE DETECTION STAGE

| CAD SYSTEM | Sensitivity (%) | Total number of candidates | Average number of candidates / scan |
|---|---|---|---|
| Murphy et al. [38] | 85.60 | 298,256 | 335.87 |
| Jacobs et al. [39] | 36.10 | 258,075 | 290.63 |
| Setio et al. [40] | 31.80 | 42,281 | 47.61 |
| Torres et al. [41] | 76.80 | 19,687 | 22.17 |
| Tan et al. [42] | 92.90 | 295,686 | 332.98 |
| Zhang et al. [18] | 100.00 | 45,939 | 51.73 |
| Setio et al. [24] | 98.30 | 754,975 | 850.20 |
| Our method | 95.36 | 18,116 | 20.40 |

In the false positive reduction stage, we assess the model in 10-fold cross-validation as well and apply the same proportion for training, validation and testing as those at the nodule candidate detection stage. The number of false positive nodules obtained from the first stage is fifteen times as many as the number of true nodules, which means the training set is imbalanced. Thus, we augment the data with different rotation degrees in three directions. With the Adam optimal function, we calculate the binary cross-entropy as the loss function. Batch size is set to 16. The learning rate starts from $10^{-4}$ since it has the best results in the preliminary experiments. We evaluate the performance by only using Archi-2 or Archi-3 to classify candidates. The performance of using two architectures is also assessed. We categorize candidates according to their predicted sizes of bounding boxes. Then we perform the false positive reduction step to cube images with either Archi-2 or Archi-3 depending on the predicted sizes. We fuse the probability by weighted linear combination and use grid search to determine ratios. The weights from Archi-2, Archi-3 and the combination of architectures are one third. The ensemble of prediction results provides complementary information, which makes final predictions more accurate and robust to differences of nodule size.

The framework is implemented in Python based on the deep learning library of Keras [36] with the TensorFlow backend.

## IV. EXPERIMENTAL RESULTS

### A. Candidates Detection

Table I shows the performance of four streams that have input images in different slab thicknesses and their merged architecture. The networks trained by 1 mm axial section slice set, 5 mm MIP set, 10 mm MIP set and 15 mm MIP set showed sensitivity of 82.80%, 87.86%, 89.97%, 88.70%, respectively. After we combined the results from four streams, the CAD system achieved sensitivity of 95.36% with 19.13 false positives (FPs) per scan. The results showed single thicker MIP images increased the performance of nodule detection compared to 1 mm axial section slices. This also demonstrated the fused framework achieved a considerable improvement on pulmonary nodule detection by using varied MIP images. We categorized nodules according to their sizes in diameter (i.e., 3-10 mm, 10-20 mm, ≥20 mm). It is worth noting that these four streams detected a similar number of nodules with the diameter larger than 10 mm and did not miss any nodules with a diameter larger than 20 mm. Among these four streams, the one trained with 10 mm MIP images detected the largest number of nodules in small sizes with a relatively low false positive rate, whereas the one trained with 1 mm axial section slices had the lowest sensitivity with more FPs. It can be also noticed that the convolutional neural network utilizing MIP images for training could acquire the improvement of sensitivity in nodule detection, especially for relatively small nodules (3-10 mm). Compared to the first stream, other streams trained by thicker MIP images have fewer number of false positives.

### B. False Positive Reduction

The free-response receiver operating characteristic (FROC) [37] curves resulting from our developed CAD system is displayed in Fig. 5. We evaluated the performance of the system at various low false positive rates at x-axis (i.e. 0.125, 0.25, 0.5, 1, 2, 4, 8 and 16 FPs/scan). The false positive rate decreases from 16 FPs/scan to 2 FPs/scan, but the sensitivity remained steady This evidence showed effectiveness of an ensemble of multiple CNNs for false positive reduction. All single architectures acquired sensitivity of over 90.75% with 1 FPs per scan. Our fusion results obtained a good sensitivity of 89.89% and 94.81% at 0.25 and 4 FPs per scan, which proved that the incorporation of spatial nodule information in various scales increases the performance of the system.

## V. DISCUSSION

We presented a novel lung nodule detection CAD system which used multiple MIP views with different slab thicknesses from thoracic CT scans. The study aimed to explore the feasibility of applying MIP images to improve the performance of lung nodule detection using convolutional neural networks. Inspired by the clinical methodology of radiologists, this study provided a new insight on automatic pulmonary nodule detection with deep learning techniques. We showed that combining the clinical screening method and CNNs is beneficial to improve the performance of nodule detection.

TABLE III
PERFORMANCE COMPARISON WITH OTHER PUBLISHED COMPUTER-AIDED DETECTION SYSTEMS

| CAD system | Year | Number of Scans | Sensitivity (%) | False positives / scan |
|---|---|---|---|---|
| Tan et al. [43] | 2013 | 360 | 83.00 | 4.0 |
| van Ginneken et al. [44] | 2015 | 865 | 76.00 | 4.0 |
| Torres et al. [41] | 2015 | 949 | 80.00 | 8.0 |
| Ypsilantis et al. [13] | 2016 | 1080 | 90.50 | 4.5 |
| Setio et al. [14] | 2016 | 888 | 85.40 | 1.0 |
|  |  |  | 90.10 | 4.0 |
| Firmino et al. [45] | 2016 | 420 | 94.40 | 7.0 |
| Dou et al. [15] | 2017 | 888 | 84.80 | 1.0 |
|  |  |  | 90.70 | 4.0 |
| Setio et al. [24] | 2017 | 888 | 96.90 | 1.0 |
|  |  |  | 98.20 | 4.0 |
| Jiang et al. [16] | 2018 | 1006 | 80.06 | 4.7 |
|  |  |  | 94.00 | 15.1 |
| Zhang et al. [18] | 2018 | 888 | 94.90 | 1.0 |
|  |  |  | 97.20 | 4.0 |
| **Our framwork** | **2019** | **888** | **92.67** | **1.0** |
|  |  |  | **94.19** | **2.0** |

Comparing to the existing methods which were accessed on the LIDC/IDRI database, our MIP-based CAD system demonstrated a strong capability in nodule detection while maintaining a low false positive rate.

Candidate detection is an essential step for CAD systems. We considered the efficiency and feasibility in different ways to combine the results from the networks for nodule detection. Among all the possible combinations, merging the results from four groups improved the performance from the lowest sensitivity of 82.80% to 95.36%, which is a significant improvement. Table II shows the performance of individual candidate detection systems in the LUNA16 competition. Murphy et al.[38], Jacobs et al. [39], Setio et al. [40], M5L Torres et al.[41], and Tan et al. [42] employed classical algorithms based on machine learning, while Zhang et al. [18], Setio et. al [24] and our proposed system applied methods based on deep learning. It is interesting to note that the results from deep learning approaches showed better sensitivities on nodule detection, which indicates the generality of deep learning. Zhang et al. [18] and Setio et. al [24] had better detection rates. However, in order to detect as many nodules as possible, these CAD systems localized more FPs. Our system had the least number of candidates per scan with sensitivity comparable to that in other systems.

To provide a comparative analysis and show the generality of these systems, we chose the approaches which had a large-scale evaluation. Our proposed method focused on the 888 scans in the whole LIDC/IDRI database which contains 1186 nodules accepted by the majority of four radiologists. We listed the recent results of lung nodule detection systems from other researchers in Table III. Although Jiang et al. [16] and Firmino et al. [45] achieved competitive sensitivities in their methods, there were still many false positives in the results. Thus, radiologists still had to make more effort to distinguish true nodules from potential candidates. Despite having a low false positive rate, the detection rates achieved by Setio et al. [14] and Dou et al. [15] needed further improvement. Setio et al. [24] had better performance on nodule detection since they included some CAD systems which were designed for special types of nodules, such as subsolid, juxta-vascular, juxta-pleural nodules, etc. Besides, they combined the results from seven complete nodule detection systems and five false positive reduction systems with varied architectures. However, our proposed method was designed for all nodules. It had comparable results with only one system which contained two types of CNNs, and required less time for processing. Zhang et al. [18] tried to include any suspicious candidates by employing multi-scale LoG filters and acquired a better result at the nodule candidate detection stage. But it had more false positives at the candidate detection stage compared to our study. Our networks with thicker MIP images could detect nodules with fewer false positives because of the ability of MIP to differentiate between nodules and vessels.

We used 2-D CNNs to localize the nodules in our framework. Although 3-D CNNs might extract more spatial information of nodules with higher discrimination capability, they demanded more computational power and the training time. In contrast, 2-D approaches required lower computing resource and less storage space, which was more applicable and efficient in practice. Benefiting from MIP images, we added extra structural 3-D spatial information to 2-D images, which enhanced the contextual discrimination between nodules and vessels and provided fewer number of potential candidates for false positive reduction stage. They can be localized more accurately after the first stage, which means that the architecture could analyze them in much smaller cubic patches with affordable computation. To remove as many false positives as possible and keep robust performance, we ensembled 3-D networks for candidate nodule classification.

One of the big challenges of deep learning methods was the acceptance of their results by the radiologists since CNN models were not very easily interpretable to them. Following the procedures of radiologists, we connected the clinical work to this task by using varied MIP images. We tried to provide an explanation by showing that CNNs could accurately find nodules based on the MIP images which were used clinically during screening practices.

Regarding the use of our proposed method for clinical lung cancer screening, there is still room for improvement on the detection rate and false positive rate. We can optimize our work in the future in these directions. It is of interest to evaluate our lung nodule detection system on more lung screen trial data and low-dose CT scans. Furthermore, we need to improve the detection performance on small nodules and ground nodules since they can be easily overlooked during the screening.

## VI. CONCLUSION

In this work, we design a novel lung nodule detection CAD system, applying multiple MIP views with various slab thicknesses. The performance of the framework demonstrates the significance and effectiveness of integrating multi-slab thicknesses MIP images for lung nodule detection in CT scans. The combination of networks in various scales yields accurate and robust performance for false positive reduction. We also prove that MIP images can provide conspicuous benefits when exploiting CNNs to detect pulmonary lesions, especially small ones. The proposed CAD system can improve the radiologists' work efficiency by largely reducing the number of scans that needs to be evaluated. Our designed system will be further explored and validated on other clinical data. Hopefully, it will be promoted in lung cancer screening.

## ACKNOWLEDGMENT

The authors would like to thank Google for providing us with a research grant to run our computations on the Google Cloud Platform and NVIDIA for the support of the GPU. We would also like to thank the LUNA16 challenge for providing dataset to this research.